\pdfoutput=1

\documentclass[11pt]{article}

\usepackage[]{acl}

\usepackage{times}
\usepackage{latexsym}

\usepackage{amsmath}
\usepackage{amssymb}
\usepackage{framed} 
\usepackage{subcaption}  
\usepackage{tabularx}  
\usepackage{booktabs}  
\usepackage{multirow}  
\usepackage{makecell}
\makeatletter
\def\new@fontshape{}
\makeatother
\usepackage{siunitx} 

\usepackage{xspace}  

\usepackage{tikz}  
\usetikzlibrary{shapes.geometric, arrows}
\usetikzlibrary{calc, positioning}  
\usepackage{forest}  
\useforestlibrary{linguistics}
\forestapplylibrarydefaults{linguistics}

\usepackage{stmaryrd} 

\usepackage[compress,sort,capitalise]{cleveref}  
\usepackage{hyperref}

\usepackage[inline]{enumitem}  

\usepackage{gb4e}  
\noautomath

\newcommand{\lform}[1]{\texttt{#1}}

\crefname{xnumi}{}{}
\creflabelformat{xnumi}{(#2#1#3)}
\crefrangeformat{xnumi}{(#3#1#4)--(#5#2#6)}

\tikzstyle{nnode} = [ellipse, text centered, draw=black, inner sep=2pt] 
\tikzstyle{dnode} = [rectangle, rounded corners, text centered, draw=black] 
\tikzstyle{arrow} = [thick,->,>=stealth]

\definecolor{colorwant}{HTML}{ABBEF1} 
\definecolor{colorgo}{HTML}{DE8BB3} 
\definecolor{colorthe}{HTML}{F1AD83} 
\definecolor{colorboy}{HTML}{FFE19D} 

\pgfdeclarelayer{background}
\pgfdeclarelayer{main}
\pgfsetlayers{background,main}

\usepackage{color}
\usepackage{bm}
\definecolor{orange}{rgb}{1,0.5,0}
\definecolor{mdgreen}{rgb}{0.05,0.6,0.05}
\definecolor{mdblue}{rgb}{0,0,0.7}
\definecolor{dkblue}{rgb}{0,0,0.5}
\definecolor{dkgray}{rgb}{0.3,0.3,0.3}
\definecolor{slate}{rgb}{0.25,0.25,0.4}
\definecolor{gray}{rgb}{0.5,0.5,0.5}
\definecolor{ltgray}{rgb}{0.7,0.7,0.7}
\definecolor{purple}{rgb}{0.7,0,1.0}
\definecolor{lavender}{rgb}{0.65,0.55,1.0}
\definecolor{brown}{rgb}{0.6,0.2,0.2}

\newcommand{\code}[1]{}

\newcommand{\meandiscrim}{\textit{Mean}\textsubscript{\textit{discrim}}}
\newcommand{\meanbounds}{\textit{Mean}\textsubscript{\textit{bounds}}}

\renewcommand{\paragraph}[1]{\textbf{#1}}
\newcolumntype{g}{>{\columncolor{orange}}c}

\usepackage[T1]{fontenc}

\usepackage[utf8]{inputenc}

\usepackage{microtype}

\usepackage{inconsolata}

\title{Predicting generalization performance with correctness discriminators}

\author{Yuekun Yao \and Alexander Koller \\
  Department of Language Science and Technology\\
  Saarland Informatics Campus\\
  Saarland University, Saarbrücken, Germany \\
  \texttt{\{ykyao, koller\}@coli.uni-saarland.de} \\}

\begin{document}
\maketitle

\begin{abstract}
    The ability to predict an NLP model's accuracy on unseen, potentially out-of-distribution data is a prerequisite for trustworthiness. We present a novel model that establishes upper and lower bounds on the accuracy, without requiring gold labels for the unseen data. We achieve this by training a \emph{discriminator} which predicts whether the output of a given sequence-to-sequence model is correct or not. We show across a variety of tagging, parsing, and semantic parsing tasks that the gold accuracy is reliably between the predicted upper and lower bounds, and that these bounds are remarkably close together.
\end{abstract}

\section{Introduction}

In order for a user to trust that an  NLP system performs its task with sufficient reliability, the user must be able to judge the system's accuracy on real-world tasks of interest. This ability is growing in importance with the rapidly increasing prominence of NLP technology in users' daily lives and the growing capability of this technology to solve high-level tasks, e.g.\ by orchestrating the use of external tools \citep{yao2023react,NEURIPS2023_1b44b878}. At the same time, even the best available models still struggle on out-of-distribution (OOD) test sets \cite{lake-baroni-2018-generalization,li2023slog} and complex tasks on unseen domains \citep{zhou2024webarena,jimenez2024swebench}.

In realistic settings, the accuracy of an NLP model $M$ needs to be estimated on \emph{unlabeled} test data; it is plausible that the estimator has access to the user's inputs, but not to gold annotations that would capture the behavior the user intended. There is some previous work on estimating the accuracy of $M$ on unlabeled test data, primarily for text or image classification tasks and based on $M$'s confidence \cite{garg2022leveraging, guillory2021predicting}. However, existing accuracy estimation models provide only point estimates for the accuracy of $M$, which hides their own uncertainty; a user cannot judge whether the accuracy estimator is confident about its estimates or whether they should be cautious about trusting them. Ultimately, there is an infinite hierarchy of accuracies: the true accuracy of $M$; the accuracy of the accuracy estimator; estimates of \emph{that} accuracy; and so on.

\begin{figure}
    \centering
    \includegraphics[scale=0.47]{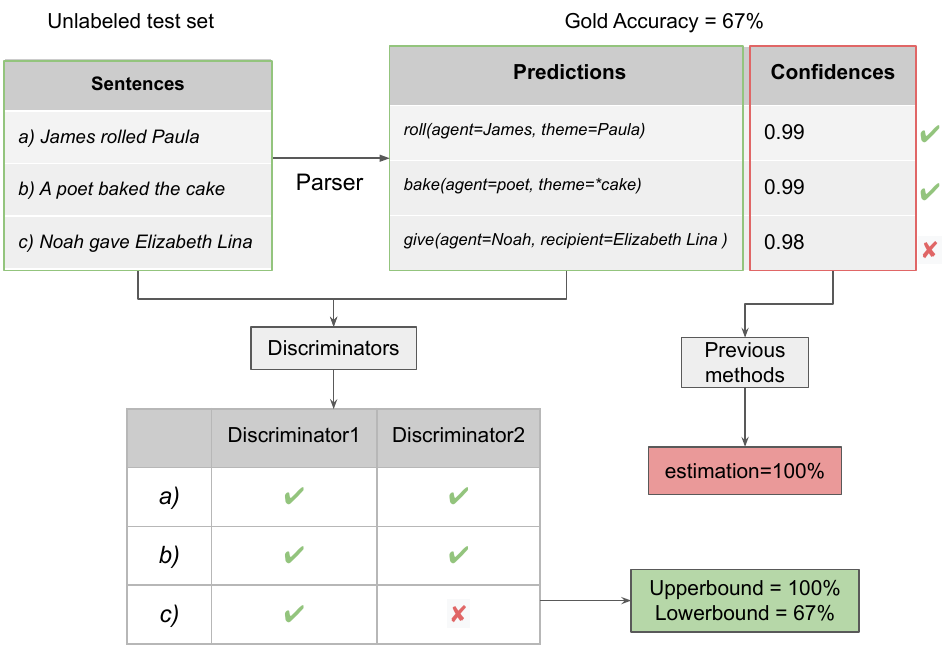}
    \caption{Comparison of our discriminators and confidence-based methods. Our method provides upper and lower bounds which can capture gold accuracy.}
    \label{fig:intro}
\end{figure}

In this paper, we take the first step up this hierarchy by offering a method for capturing the accuracy predictor's uncertainty about the estimation of $M$'s accuracy. Instead of directly calculating a point estimate for $M$'s accuracy, our method predicts  \emph{upper and lower bounds} for this accuracy, from unlabeled test data. We focus on estimating the accuracy of sequence-to-sequence models, applied to parsing, semantic parsing, and tagging tasks; these tasks have the advantage over other sequence generation tasks that there is a unique correct answer, which allows us to talk about accuracies. 

We first train a \emph{discriminator} to predict whether $M$'s output on a given input is correct or not; we show that this can be done with remarkable accuracy across a range of tasks. We then run an ensemble of discriminators on $M$'s predictions on the unlabeled test data and obtain upper and lower bounds through a voting mechanism (Figure \ref{fig:intro}). We show across a variety of in-distribution and OOD tasks that $M$'s true accuracy is reliably between the upper and lower bounds, and that these bounds are quite tight. Finally, if forced to predict point estimates of the true accuracy, our model provides more precise estimates than earlier work on most datasets, by taking the mean of the upper and lower bounds.

We will release our code online \footnote{\url{https://github.com/coli-saar/discriminator}}.

\section{Related work}

\paragraph{Calibration.}
A neural model is called \textit{well-calibrated} if its predicted probability (e.g.\ confidence) for its decision (e.g.\ label or sequence) aligns to the probability of the prediction correctness. 
Much prior work has attempted to improve the calibration of systems,
often through either modifying training objectives or posthoc methods: \citet{kong-etal-2020-calibrated} add a regularization term into training objective to address in-distribution calibration and out-of-distribution detection for text classification; \citet{desai-durrett-2020-calibration} exploit temperature scaling \cite{guo2017calibration} to normalize output logits with a scalar temperature parameter; \citet{dong-etal-2018-confidence, kamath-etal-2020-selective} train an additional regressor to estimate the model confidence with designed features for semantic parsing; \citet{jiang-etal-2021-know} investigate all these methods and find that posthoc-based methods are universally helpful for question answering tasks.

Most calibration works above focus on in-distribution (ID) tasks and assume a development set as given, which allows them to estimate parameters (e.g.\ temperature) to yield the optimal confidence. 
However, according to \citet{kamath-etal-2020-selective}, the predicted model confidence is an unreliable estimate of the correctness on OOD generalization tasks.
Compared to such calibration works, 
our method applies just as easily to OOD as to ID tasks.
Further, development sets from OOD distributions are usually difficult to access, which introduces additional challenges of applying calibration-based methods.
\citet{kamath-etal-2020-selective} also consider distribution shift, but their calibrator requires a small amount of data from a known OOD distribution.


\paragraph{Predicting test accuracy from unlabeled data.}
Previous works have investigated predicting the model performance on an unannotated OOD test set for other tasks: \citet{guillory2021predicting} exploit the difference of confidences between training distribution and the OOD distribution as a useful feature; \citet{jiang2021assessing} show that the test error of deep networks can be estimated by the disagreement of two models trained with the same architecture on the same training set but with two different runs; \citet{yu2022predicting} exploits the euclidean distance between model parameters trained on differently distributed data to predict generalization errors; \citet{garg2022leveraging} estimate a threshold of model confidence from training data and predict the correctness of OOD data based on it; \citet{fu2023estimating} train an additional model to predict the accuracy of large language models on question answering tasks, which takes as input confidence scores and outputs the overall accuracy of the test set.  


Works introduced above estimate the accuracy as a scalar value between 0 and 1.
In contrast, our method explicitly judges the uncertainty of the estimated accuracy, providing upper and lower bounds for the estimated accuracy. 
Besides, previous works only consider image classification and natural language inference tasks.
Our work shows that for sequence generation tasks like semantic parsing, the predicted sequence can serve as a good-enough feature to determine the prediction correctness on OOD data.  

\paragraph{Quality estimation in NLP tasks.} 
Automatic accuracy prediction has also been investigated for NLP tasks: \citet{van-asch-daelemans-2010-using} exploit similarity metrics between the training and test set to estimate POS tagger performances; \citet{chatterjee-etal-2018-combining} train regressors to predict BLEU \cite{papineni-etal-2002-bleu} scores of a machine translation system with given features; \citet{opitz-frank-2019-automatic} train regressors to predict F1 scores for subtasks of AMR \cite{banarescu-etal-2013-abstract}. Compared to these works, our method does not require manually designed features and thus is easy to be adapted to any sequence generation tasks. \citet{varshney-baral-2023-post} also train a correctness discriminator to improve the coverage of a selective prediction system for question answering. In contrast to this work, we use discriminators to predict accuracies, and more specifically upper and lower bounds.

\section{Correctness discriminator}

The core of our approach is to construct and train a \emph{correctness discriminator} model, which judges the correctness of a model prediction on unseen data. In this section, we first introduce how we design the discriminator model and collect training data (Section \ref{sec:method:design}), and then describe how to predict the upper bound and lower bounds accuracy (Section \ref{sec:method:bound}). To avoid confusion, we call the model for the original parsing or tagging tasks a \textit{parser} and the model for predicting the parser performance a \textit{discriminator}. Note that here we only assume that the parser solves a sequence-to-sequence task, but the task output can be any sequence -- not just a linearized parse.

\subsection{Discriminator design} \label{sec:method:design}

The discriminator is designed as a binary classifier whose task is to determine whether a given predicted sequence is the correct output for a given natural language sentence. Formally, given a natural language sentence $X \in \mathcal{X}$ and a predicted symbolic sequence (e.g.\ meaning representation for semantic parsing tasks) $Y \in \mathcal{Y}$, the discriminator $F: \mathcal{X}\times \mathcal{Y} \to \{Correct, Incorrect\}$ maps them to a \textit{Correct} or \textit{Incorrect} label to represent its correctness. 

\begin{figure}
    \centering
    \includegraphics[scale=0.4]{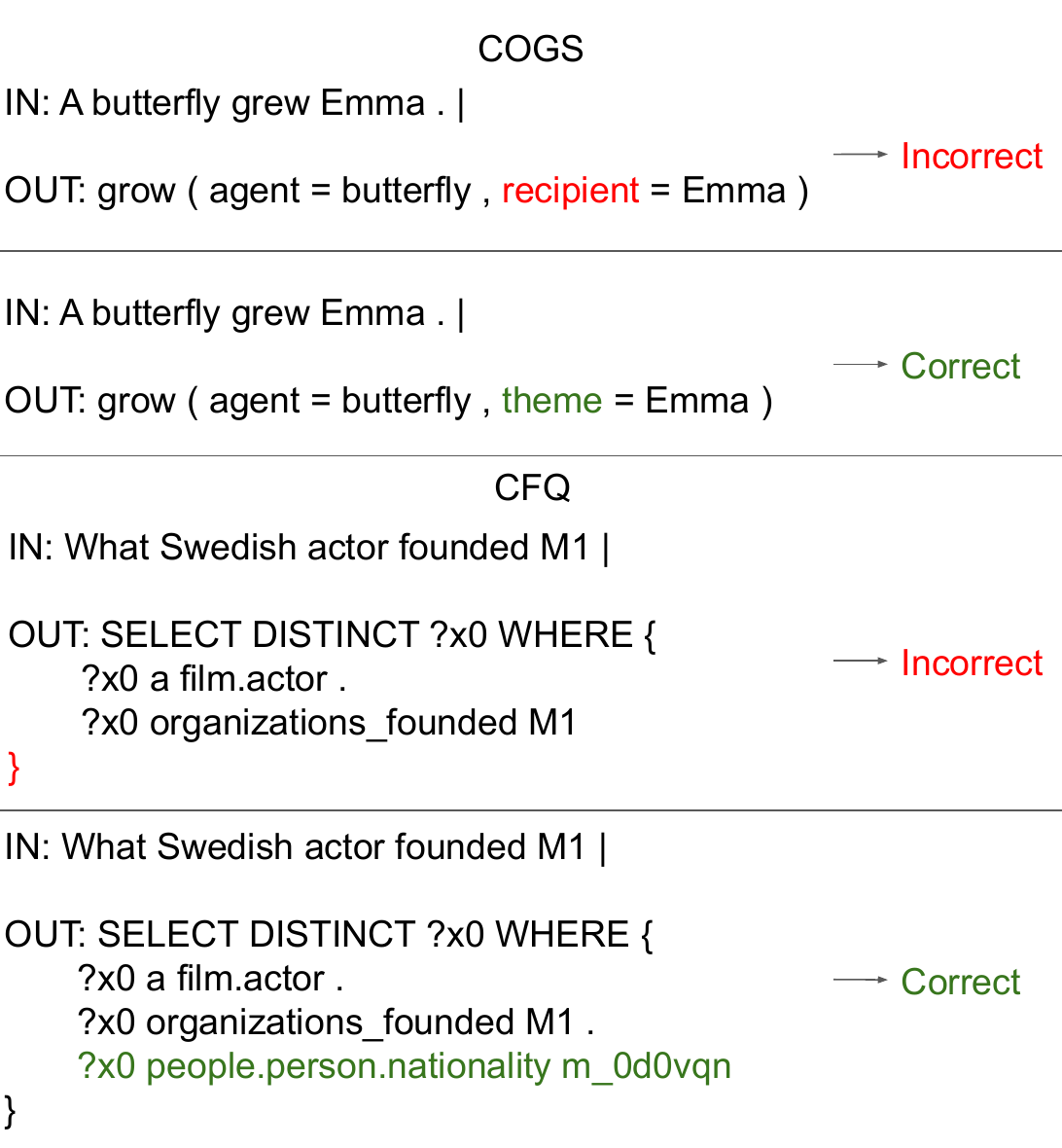}
    \caption{Examples of COGS and CFQ training data for the discriminator. \textit{IN} refers to the input sentence and \textit{OUT} refers to the predicted output sequence (e.g.\ logical form for COGS and SPARQL query for CFQ).}
    \label{fig:method:data_example}
\end{figure}

In this paper, we explore different model architectures as the discriminators: encoder-only, encoder-decoder, and decoder-only. 
All three models take as input the concatenation of the input natural language sentence with the predicted sequence.
For each architecture, we finetune an existing pretrained language model as the discriminator. 
For encoder-only (e.g.\ Roberta) discriminators, the input is first encoded into hidden representations and then fed into an additional multi-layer perceptron classifier, which determines the \textit{Correct} or \textit{Incorrect} label. 
For encoder-decoder (e.g.\ T5) and decoder-only (e.g.\ LLaMA) discriminators, 
the decoder directly generates the label.


Now we discuss how to collect training data for our discriminator.
In principle, the training data should contain both positive and negative examples. 
For positive examples, we can always exploit the training set of the parser. 
However, it is non-trivial to obtain negative examples.
Such examples can be synthesized by applying noise functions (e.g.\ replacement or deletion) to positive examples \cite{kim-etal-2021-improving}, but this requires prior knowledge about errors a parser tends to make. 
Another option is to collect errors a trained parser made on its training set, which is still challenging since parsers yield near-perfect accuracies on their training sets.

We therefore generate negative examples from intermediate checkpoints of our parser during its training.
Specifically, we run the parser checkpoint on its training data. We take incorrect predictions from the decoder beam as the negative training data for the discriminator.
Figure \ref{fig:method:data_example} gives examples of our training data.

Given the described discriminator, we can estimate the accuracy of our parser on any unseen test sets. Assuming a parser makes predictions on $|D_t|$ instances and the discriminator labels $|D_{c}|$ of predictions as \textit{Correct}, the predicted accuracy can be calculated as Eq \ref{eq:acc_pred}.
\begin{equation} \label{eq:acc_pred}
    Acc_{pred} = \frac{|D_c|}{|D_t|}
\end{equation}

\subsection{Bounds prediction} \label{sec:method:bound}

\begin{figure}
    \centering
    \includegraphics[scale=0.4]{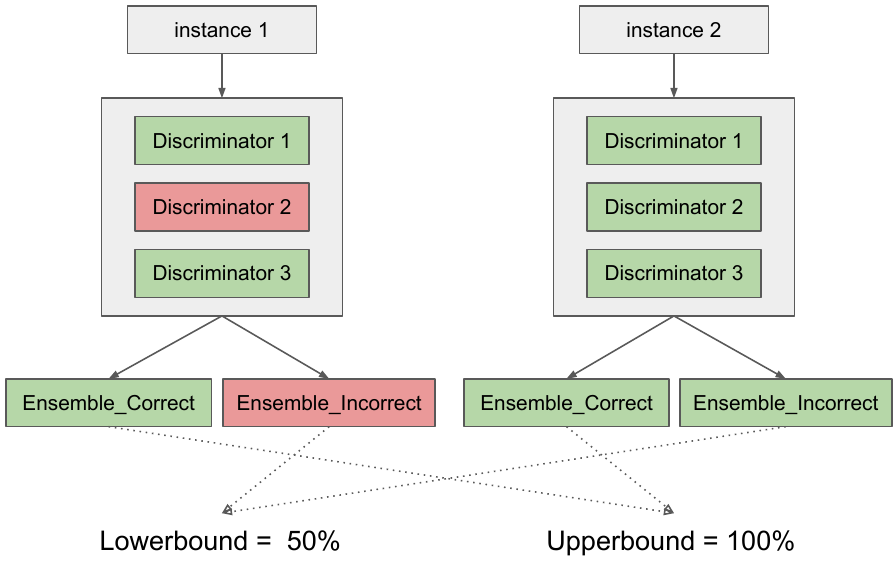}
    \caption{Example of calculating upper and lower bounds from discriminators. 
    \textit{Green} blocks mean the instance is predicted as \textit{Correct} by the discriminator and \textit{Red} blocks refer to \textit{Incorrect}. 
    \textit{Ensemble\_incorrect} predicts \textit{Incorrect} for instance 1 and \textit{Correct} for instance 2. Hence, the lower bound is $1 / 2 = 50\%$.
    }
    \label{fig:method:ensemble}
\end{figure}

We now introduce how to predict upper and lower bounds for accuracy with the discriminator described above.
This is implemented by two voting mechanisms, \textit{Ensemble\_correct} and \textit{Ensemble\_incorrect}.
These mechanisms aggregate outputs from multiple trained discriminators.
\begin{itemize}
    \item \textit{Ensemble\_correct} predicts \textit{Correct} if at least one discriminator predicts \textit{Correct}; if all discriminators predict \textit{Incorrect}, it also predicts \textit{Incorrect}.
    This yields the most optimistic estimation, assuming correctness if at least one discriminator agrees.
    \item Conversely, \textit{Ensemble\_incorrect} predicts \textit{Incorrect} if at least one discriminator predicts \textit{Incorrect}; otherwise \textit{Correct}.
    This mechanism is more cautious, predicting an instance as incorrect if any discriminator disagrees.
\end{itemize}
We calculate the upper bound of the accuracy as $Acc_{pred}$ in Eq.~\ref{eq:acc_pred}, with $|D_c|$ determined by the output of the \textit{Ensemble\_correct} mechanism instead of a single discriminator.
Similarly, the lower bound is calculated with $|D_c|$ determined by the output of \textit{Ensemble\_incorrect}.
Figure \ref{fig:method:ensemble} illustrates how the ensembles compute bounds.

Ensembles of neural networks have been shown effective for uncertainty quantification \cite{lakshminarayanan2017simple,lukovnikov-etal-2021-detecting-compositionally} by averaging confidence scores of individual models.
Here we use voting mechanisms to calculate the upper and lower accuracy bounds.

\section{Experiments} \label{sec:experiments}

We introduce our datasets, model setup, evaluation metrics and experimental results in this section.

\subsection{Datasets} \label{sec:datasets}
We experiment with three tasks: semantic parsing, part-of-speech tagging and constituency parsing.

For semantic parsing, we consider two OOD generalization scenarios: compositional generalization and low-resource domain adaptation. 
We use the COGS \cite{kim-linzen-2020-cogs} and CFQ \cite{keysers-etal-2020-measuring} datasets to evaluate compositional generalization.
For CFQ, we use its MCD1 and MCD2 splits.
For low-resource domain adaptation, we use the TOPv2 \cite{chen-etal-2020-low} dataset.
We also evaluate our method on in-distribution task with the AMR 2.0 dataset \cite{banarescu-etal-2013-abstract}.

For part-of-speech (POS) tagging and constituency parsing tasks, we use the Penn Treebank 3 (PTB) dataset \cite{marcus-etal-1993-building}. 
We train our parser on the WSJ training set and evaluate its in-domain performance on the WSJ test set and cross-domain performance on the Brown corpus. 
We predict the generalization performance for both the WSJ test set (i.e.\ in-distribution test set) and the Brown corpus (i.e.\ OOD test set), which we call Syn-WSJ, Syn-Brown (for parsing) and POS-WSJ, POS-Brown (for tagging) in this paper. 
In addition, we experiment with POS-COGS, a POS tagging dataset generated based on COGS \cite{yao-koller-2022-structural}, to evaluate compositional generalization in the POS tagging task. 
Details of our datasets are in Appendix \ref{appendix:dataset}.

\subsection{Setup}

\paragraph{Parser}. We finetune \textit{T5-base} \cite{raffel-etal-2020-t5} as the parser for all tasks described above. To do this, we convert all of our tasks into sequence generation tasks, where the output sequence can be a semantic meaning representation, POS tag sequence or linearized parse tree. 
All our parsers achieve the same or close performance as those reported in previous works using T5. 

\paragraph{Discriminator}. 
We experiment with three architectures as discriminators: (1) An encoder-only architecture consisting of a \textit{Roberta-base} encoder \cite{DBLP:journals/corr/abs-1907-11692} and an MLP classifier (2) An encoder-decoder architecture using \textit{T5-base} \cite{raffel-etal-2020-t5} and (3) A decoder-only architecture using \textit{Vicuna-7B} \cite{zheng2023judging}. 
We report results of the T5 and RoBERTa discriminators in Section \ref{sec:results}, since the T5 discriminators share the same architecture as our parsers, and the RoBERTa discriminators have the fewest parameters.
Results for the {Vicuna discriminators} in Appendix \ref{appendix:architecture}.
All three discriminators perform well across corpora.

To collect negative training examples, we validate the parser checkpoint every $K$ steps on its training set, where $K$ is a hyperparameter. Since our parser is an encoder-decoder model, we randomly sample incorrect predictions from the decoded beam predictions. 
For each task we train an ensemble of 5 discriminators with different random seeds.
See Appendix \ref{appendix:training} for more training details.

\begin{table*}[tb!]\centering
\footnotesize
\begin{tabularx}{\linewidth}{p{0.45cm}p{1.55cm}XXXXXXXXXXXXX}
\toprule
& &\multicolumn{6}{c}{T5} &\multicolumn{6}{c}{Roberta} \\
\cmidrule(lr){3-8}
\cmidrule(lr){9-14}
& &\multicolumn{2}{c}{Single} &\multicolumn{2}{c}{Upper} &\multicolumn{2}{c}{Lower} &\multicolumn{2}{c}{Single} &\multicolumn{2}{c}{Upper} &\multicolumn{2}{c}{Lower} \\
\cmidrule(lr){3-4}
\cmidrule(lr){5-6}
\cmidrule(lr){7-8}
\cmidrule(lr){9-10}
\cmidrule(lr){11-12}
\cmidrule(lr){13-14}
& &CR &IR &CR &IR &CR &IR &CR &IR &CR &IR &CR &IR \\
\midrule
\multirow{7}{*}{OOD} &MCD1 & 97.0 &83.1 &99.2 &69.8 &92.9 &94.5 & 97.0 &90.0 &98.5 &80.8 &92.5 &97.3 \\
&MCD2 & 80.6 &83.7 &83.4 &77.9 &71.7 &92.2 & 78.3 &85.0 &84.7 &79.4 &70.6 &90.6 \\
&COGS &98.5 &96.6 &99.8 &89.4 &98.5 &96.9 & 98.9 &87.7 &99.8 &86.6 &97.8 &90.8 \\
&TOP &87.4 &57.9 &92.2 &44.7 &82.5 &78.9 &85.4 &65.8 &91.3 &42.1 &77.7 &78.9 \\
&POS-Brown &81.3 &54.7 &94.0 &26.0 &52.2 &84.7 & 62.1 &67.4 &95.2 &23.3 &44.4 &88.0 \\
&POS-COGS &98.8 &86.3 &99.9 &84.4 &98.7 &89.2 &99.9 &90.9 &100 &86.7 &99.5 &94.9 \\
&Syn-Brown &33.8 &90.0 &69.3 &73.9 &29.7 &96.0 & 17.5 &74.3 &70.6 &38.4 &4.2 &96.3\\
\midrule
\multirow{3}{*}{ID} &AMR 2.0 &37.0 &98.3 &70.9 &84.1 &29.1 &99.1  &38.9 &96.1 &59.1 &80.2 &37.3 &97.4 \\
&POS-WSJ &80.2 &53.6 &93.0 &26.5 &52.4 &84.8 &64.5 &65.2 &96.0 &21.7 &49.6 &87.6  \\
&Syn-WSJ &44.5 &89.2 &66.2 &73.1 &20.5 &97.0 & 27.1 &66.0 &74.3 &34.6 &6.5 &94.7 \\
\bottomrule
\end{tabularx}
\caption{Results of our discriminators on different datasets. 
    For each dataset, we report \textit{Correct-Recall} (CR) and \textit{Incorrect-Recall} (IR). 
    \textit{Single} refers to the results with predictions from a single discriminator.
    \textit{Upper} refers to the results with discriminator predictions using \textit{ensemble\_incorrect};
    \textit{Lower} to \textit{ensemble\_correct}.
}
\label{tab:discrim:single_results}
\end{table*}

\paragraph{Comparable baseline}. We also compare our methods with several previous methods.

\textit{MaxProb}. Maxprob is a strong baseline shown in \citet{kamath-etal-2020-selective}. Assuming we are given a threshold $\gamma$ on the maximal prediction probability (e.g.\ confidence) of a parser, we can predict an instance as \textit{Correct} if the parser confidence on this instance is higher than $\gamma$, otherwise \textit{Incorrect}. Since we have no prior knowledge about the OOD distribution, we set $\gamma=0.5$ in our experiments.

\textit{Average Confidence (AC)}. We take the average confidence across the test set as the predicted accuracy. Different from previous works where the confidence is defined as the maximal softmax probability of the classifier, here we define the confidence as the probability of the most probable sequence in the beam, which is calculated by the product of softmax probabilities of each word in the sequence. 

\textit{Difference Of Confidence (DOC)}.
We also estimate the accuracy using DOC \cite{guillory2021predicting}. We start with a development set that follows the same distribution as the training data. We then subtract the difference in average confidence between the development set and the test set from the gold accuracy on the development set. 
The result is the estimated accuracy on the test set.

\textit{Average Thresholded Confidence (ATC)} 
is a strong method recently proposed by \citet{garg2022leveraging}, which has been shown to be more effective than previous methods. Applying ATC consists of two steps. First, we estimate a threshold $\gamma$ on parser confidence scores to make the number of errors made by the parser match the number of instances where the parser confidence is lower than $\gamma$; then we can obtain the predicted accuracy on the test set by calculating the fraction of unlabeled instances that obtain a score below $\gamma$.

\textit{Maxprob (Oracle)}. To compare with our predicted bounds, we calculate bounds based on Maxprob, where 
estimate $\gamma$ such that the \textit{Correct-Recall} calculated based on $\gamma$ is equal to the one from the predicted upper bound calculated by our discriminators. 
This measures the reliability of the parser's confidence in recognizing correct instances compared to discriminators.
Similarly, we calculate a lower bound by matching \textit{Incorrect-Recall} scores. 
Note that this method requires annotated test sets, which are impractical for real-world applications.



\begin{table*}[tb!]\centering
\footnotesize
\begin{tabularx}{\linewidth}{lXXXXXXXXXX}
\toprule
& \multicolumn{8}{c}{OOD} & \multicolumn{2}{c}{ID} \\
\cmidrule(lr){2-9}
\cmidrule(lr){10-11}
&\multicolumn{2}{c}{MCD1} &\multicolumn{2}{c}{MCD2} &\multicolumn{2}{c}{COGS} &\multicolumn{2}{c}{TOP} &\multicolumn{2}{c}{AMR 2.0} \\\cmidrule{2-11}
&Acc &AE $\downarrow$ &Acc &AE $\downarrow$&Acc &AE $\downarrow$&Acc &AE $\downarrow$ &Acc &AE $\downarrow$\\\midrule
\textit{Maxprob} &84.5 &26.7 &78.0 &55.1 &97.1 &5.7 &92.2 &19.2 &40.6 &26.3 \\
\textit{AC} &82.5 &24.7 &74.0 &51.1 &96.6 &5.2 &85.9 &12.9 &38.0 &23.7 \\
\textit{DOC} &82.9 &25.1 &74.3 &51.4 &96.6 &5.2 &89.3 &16.3 &32.8 &18.5 \\
\textit{ATC} &73.0 &15.2 &56.9 &34.0 &100.0 &8.6 &66.0 &7.0 &15.0 & {0.7} \\
\midrule
\multicolumn{11}{l}{\textit{Maxprob (Oracle)}} \\
\midrule
Upper. &86.4 &- &51.3 &- &96.7 &- &85.8 &- &17.9 &- \\
Lower. &43.7 &- &17.2 &- &44.3 &- &65.2 &- &5.8 &- \\
Mean &65.1 &7.3 &34.3 &11.4 &70.5 &20.9 &75.5 &2.5 &11.8 &2.5 \\
\midrule

\multicolumn{11}{l}{\textit{Ours (T5)}} \\
\midrule
{Mean}\textsubscript{discrim} & 63.0 &5.2 &28.8 &\textbf{5.9} &91.4 &\textbf{0.0} &75.3 &2.3 &8.6 &5.7 \\
Upper. &\textcolor{mdgreen}{70.0} &- &\textcolor{mdgreen}{36.1} &- &\textcolor{mdgreen}{92.1} &- &\textcolor{mdgreen}{82.3} &- &\textcolor{mdgreen}{18.3} &- \\
Lower. &\textcolor{mdgreen}{56.0} &- &\textcolor{mdgreen}{22.4} &- &\textcolor{mdgreen}{90.2} &- &\textcolor{mdgreen}{66.0} &- &\textcolor{mdgreen}{3.4} &- \\
{Mean}\textsubscript{bounds} &63.0 &{5.2} &29.3 &{6.4} &91.2 &{0.3} &74.2 &{1.2} &10.9 &{3.4} \\
\midrule
\multicolumn{11}{l}{\textit{Ours (Roberta)}} \\
\midrule
{Mean}\textsubscript{discrim} & 59.5 &\textbf{1.7} &28.9 &6.0 &91.5 &0.1 &73.0 &\textbf{0.0} &14.1 &\textbf{0.2} \\
Upper. &\textcolor{mdgreen}{65.0} &- &\textcolor{mdgreen}{35.3} &- &\textcolor{mdgreen}{92.3} &- &\textcolor{mdgreen}{82.3} &- & \textcolor{mdgreen}{25.5}& -\\
Lower. &\textcolor{mdgreen}{54.6} &- &\textcolor{red}{23.4} &- &\textcolor{mdgreen}{90.2} &- &\textcolor{mdgreen}{62.4} &- & \textcolor{mdgreen}{7.6}& -\\
{Mean}\textsubscript{bounds} &59.8 &{2.0} &29.3 &{6.4} &91.3 &{0.2} &72.3 &{0.7} & 16.6& 2.3\\
\midrule

Gold &57.8 &0.0 &22.9 &0.0 &91.4 &0.0 &73.0 &0.0 &14.3 &0.0 \\
\bottomrule
\end{tabularx}
\caption{Predicted test-set accuracy on semantic parsing tasks. 
    \textit{Upper.} and \textit{Lower.} in the leftmost column refer to predicted upper bound and lower bound. 
    \meandiscrim\ refers to a point estimate obtained as \textit{the mean of estimated accuracies by all discriminators}.
    \meanbounds\ refers to \textit{the mean of estimated upper and lower bounds} as the point estimate.
    \textit{Gold} refers to the accuracy evaluated with gold annotations. 
    \textcolor{mdgreen}{\textit{Green}} numbers refer to valid bounds that capture the gold accuracy, and  \textcolor{red}{\textit{Red}} numbers refer to invalid bounds.
}
\label{tab:discrim:sem_acc}
\end{table*}

\subsection{Evaluation metrics}
For all parsing tasks, we evaluate the exact match accuracy of our parser. 

For discriminators, we need a metric to quantify the quality of the predicted upper and lower bounds. Intuitively, such a metric should reflect \textit{whether the gold accuracy is within the bounds} (i.e.\ reliability) and \textit{whether the bounds are tight} (i.e.\ tightness). 

Previous work predicts point estimations for OOD test sets and evaluates their method with mean absolute estimation error (MAE) by calculating average absolute difference between the true accuracy on the target data and the estimated accuracy on the same unlabeled examples. 
Their results are averaged over multiple test sets for each classifier (e.g.\ parser in our tasks). 
In our setup, most tasks only have one OOD or ID test set, and thus we calculate the absolute estimation error (AE) to compare with previous works. 
Equation \ref{eq:ae} defines the metric, where $Acc_{gold}$ denotes the gold accuracy and $Acc_{pred}$ denotes the predicted accuracy. 
\begin{equation}
    |Acc_{gold} - Acc_{pred}|
    \label{eq:ae}
\end{equation}
We calculate $Acc_{pred}$ with two methods: (1) calculating the mean of estimated accuracies by all discriminators and (2) calculating the mean of estimated upper and lower bounds.
Despite their simplicity, both methods perform well across tasks.

In addition, we report the \textit{Recall} of our discriminators. Specifically, we report the score for the \textit{Correct} and \textit{Incorrect} labels individually.
We define \textit{True Correct (TC)} as instances with an annotation being \textit{Correct} and the prediction being \textit{Correct}, \textit{False Correct (FC)} as instances with an annotation being \textit{incorrect} and the prediction being \textit{correct}. Similarly, we can define \textit{True Incorrect} (TI) and \textit{False Incorrect} (FI). The \textit{Correct-Recall} is calculated by Equation \ref{eq:cr}; the \textit{Incorrect-Recall} is analogous.
\begin{equation}
    CR = \frac{Count(TC)}{Count(TC)+Count(FI)} 
    \label{eq:cr}
\end{equation}

These recall scores indicate how many correct or incorrect instances can be discriminated, but are not studied by previous works. We propose these metrics as a side contribution, which can be beneficial for downstream uses of the discriminator. 


\subsection{Results} \label{sec:results}


\paragraph{Correctness of bounds prediction.} 
We first report recall scores of our discriminators in Table \ref{tab:discrim:single_results}.
For both T5 and RoBERTa discriminators, we can observe that the upper bound achieves the highest \textit{Correct-Recall} score, and the lower bound achieves the highest \textit{Incorrect-Recall} score. 
This is because these bounds are based on voting mechanisms specifically designed to find correct or incorrect predictions.
On many of our datasets, these recall scores approach 100\%, which indicates the strong ability of our method to discriminate correctness.   

\paragraph{Accuracy of bounds prediction.}
We then compare the predicted accuracy of our bounds in Table \ref{tab:discrim:sem_acc} (e.g.\ semantic parsing), Table \ref{tab:discrim:pos_acc} (e.g.\ tagging) and Table \ref{tab:discrim:syn_acc} (e.g.\ parsing). 
We can observe that our predicted upper and lower bounds accurately capture the gold accuracy (i.e.\ high reliability). 
This pattern holds for 9 of 10 datasets with {T5} discriminators, and for 8 of 10 datasets with RoBERTa discriminators. 
Even for POS-COGS and MCD2, where this conclusion is not true, the gold accuracy only violates the bounds predicted by a small amount (i.e.\ 0.4\% on POS-COGS and 0.5\% on MCD2). 
Meanwhile, the predicted upper and lower bounds are usually close (i.e.\ high tightness). 
Comparing our predicted bounds with \textit{Maxprob (Oracle)}, our bounds are more tight on OOD generalization tasks (e.g.\ MCD splits and COGS).
Note that \textit{Maxprob (Oracle)} can access gold annotations to find a proper bound, which is implausible in practice.
Nonetheless, our method still provides better bounds than this oracle method, indicating the effectiveness of our method on OOD tasks.


\paragraph{Obtaining point estimates.}
We also compare our method with other point estimation methods with two heuristics (e.g.\ \meandiscrim\ and \meanbounds\ rows in \textit{Ours}). 
Although our methods are not specifically designed for point estimation, these estimates substantially outperform previous methods and achieve very low AE scores across all tasks.
Our method is especially useful for OOD test sets, where confidence-based methods yield a much larger AE. 


\begin{table}[tb!]\centering
\footnotesize
\begin{tabularx}{\linewidth}{p{1.3cm}XXXXXX}\toprule
& \multicolumn{4}{c}{OOD} & \multicolumn{2}{c}{ID} \\
\cmidrule(lr){2-5}
\cmidrule(lr){6-7}
&\multicolumn{2}{c}{POS-Brown} &\multicolumn{2}{c}{POS-COGS} &\multicolumn{2}{c}{POS-WSJ} \\\cmidrule{2-7}
&Acc &AE  &Acc &AE &Acc &AE \\
\midrule
\textit{Maxprob} &87.4 &26.4 &99.8 &14.1 &84.7 &19.4 \\

\textit{AC} &80.5 &19.5 &100.0 &14.3 &77.4 &12.1 \\
\textit{DOC} &88.0 &27.0 &98.8 &13.1 &85.0 &19.7 \\
\textit{ATC} &68.0 &7.0 &100.0 &14.3 &61.6 &3.7 \\
\midrule
\multicolumn{7}{l}{\textit{Maxprob (Oracle)}} \\\midrule
Upper. &83.6 &- &99.6 &- &82.4 &- \\
Lower. &44.5 &- &83.3 &- &47.9 &- \\
Mean &64.0 &3.0 &91.4 &5.7 &65.2 &0.1 \\\midrule
\multicolumn{7}{l}{\textit{Ours (T5)}} \\\midrule
{Mean}\textsubscript{discrim} &64.1 &3.1 &87.1 &1.4 &65.6 &\textbf{0.3} \\
Upper. &\textcolor{mdgreen}{86.2} &- &\textcolor{mdgreen}{87.9} &- &\textcolor{mdgreen}{86.2} &- \\
Lower. &\textcolor{mdgreen}{37.8} &- &\textcolor{red}{86.2} &- &\textcolor{mdgreen}{39.5} &- \\
{Mean}\textsubscript{bounds} &62.0 &\textbf{1.0} &87.1 &{1.4} &62.9 &{2.4} \\\midrule
\multicolumn{7}{l}{\textit{Ours (Roberta)}} \\\midrule
{Mean}\textsubscript{discrim} & 62.6 &1.6 &86.8 &1.1 &65.8 &0.5 \\
Upper. &\textcolor{mdgreen}{88.0} &- &\textcolor{mdgreen}{87.6} &- &\textcolor{mdgreen}{89.9} &- \\
Lower. &\textcolor{mdgreen}{31.8} &- &\textcolor{red}{86.1} &- &\textcolor{mdgreen}{36.7} &- \\
{Mean}\textsubscript{bounds} &59.9 &{1.1} &86.8 &\textbf{1.1} &63.3 &{2.0} \\
\midrule
Gold &61.0 &0.0 &85.7 &0.0 &65.3 &0.0 \\
\bottomrule
\end{tabularx}
\caption{Predicted accuracy on POS tagging tasks.}
\label{tab:discrim:pos_acc}
\end{table}

\begin{table}[tb!]\centering
\footnotesize
\begin{tabularx}{\linewidth}{lXXXX}\toprule
&\multicolumn{2}{c}{OOD} &\multicolumn{2}{c}{ID} \\
\cmidrule(lr){2-3}
\cmidrule(lr){4-5}
&\multicolumn{2}{c}{Syn-Brown} &\multicolumn{2}{c}{Syn-WSJ} \\\cmidrule{2-5}
&Acc &AE &Acc &AE  \\\midrule
\textit{Maxprob} &48.3 &14.5 &50.8 &13.2 \\ 
\textit{AC} &50.8 &17.0 &52.4 &14.8 \\
\textit{DOC} &48.5 &14.7 &50.2 &12.6 \\
\textit{ATC} &34.7 &{0.9} &34.0 & {3.6}\\
\midrule
\multicolumn{5}{l}{\textit{Maxprob (Oracle)}} \\\midrule
Upper. &32.9 &- &33.4 &- \\
Lower. &17.7 &- &16.5 &- \\
Mean &25.3 &8.5 &24.9 &12.7 \\\midrule
\multicolumn{5}{l}{\textit{Ours (T5)}} \\\midrule
{Mean}\textsubscript{discrim} & 36.3 &2.5 &43.8 &6.2 \\
Upper. &\textcolor{mdgreen}{57.5} &- &\textcolor{mdgreen}{64.2} &- \\
Lower. &\textcolor{mdgreen}{17.9} &- &\textcolor{mdgreen}{24.6} &- \\
{Mean}\textsubscript{bounds} &37.7 &3.9 &44.4 &6.8 \\\midrule
\multicolumn{5}{l}{\textit{Ours (Roberta)}} \\\midrule
{Mean}\textsubscript{discrim} & 40.2 &6.4 &34.8 &2.8 \\
Upper. &\textcolor{mdgreen}{64.6} &- &\textcolor{mdgreen}{68.8} &- \\
Lower. &\textcolor{mdgreen}{3.8} &- &\textcolor{mdgreen}{5.8} &- \\
{Mean}\textsubscript{bounds} &34.2 &\textbf{0.4} &37.3 &\textbf{0.3} \\
\midrule
Gold &33.8 &0.0 &37.6 &0.0 \\
\bottomrule
\end{tabularx}
\caption{Predicted accuracy on parsing tasks.}
\label{tab:discrim:syn_acc}
\end{table}

\begin{figure*}[!htb]
    \centering
    \begin{subfigure}{.3\textwidth}
        \centering
        \includegraphics[width=\linewidth]{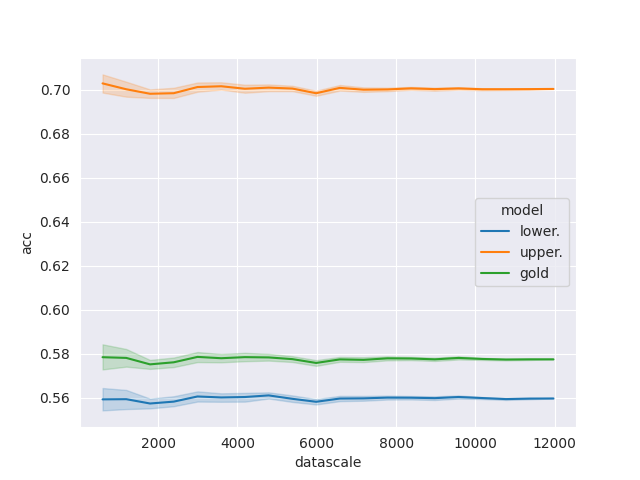}
        \caption{MCD1}
        \label{fig:subfig_a}
    \end{subfigure}%
    \begin{subfigure}{.3\textwidth}
        \centering
        \includegraphics[width=\linewidth]{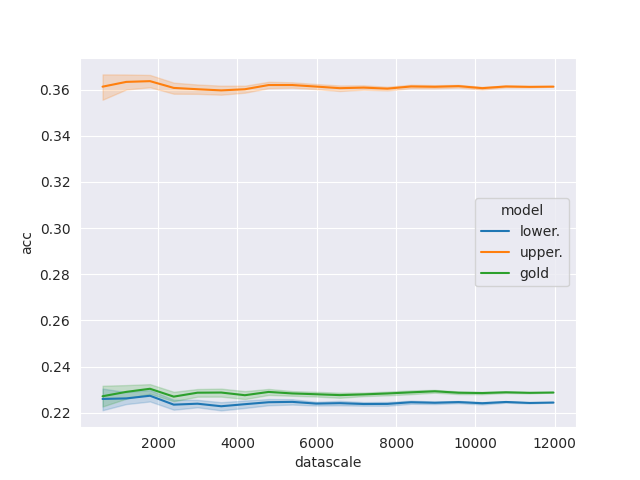}
        \caption{MCD2}
        \label{fig:subfig_b}
    \end{subfigure}%
    \begin{subfigure}{.3\textwidth}
        \centering
        \includegraphics[width=\linewidth]{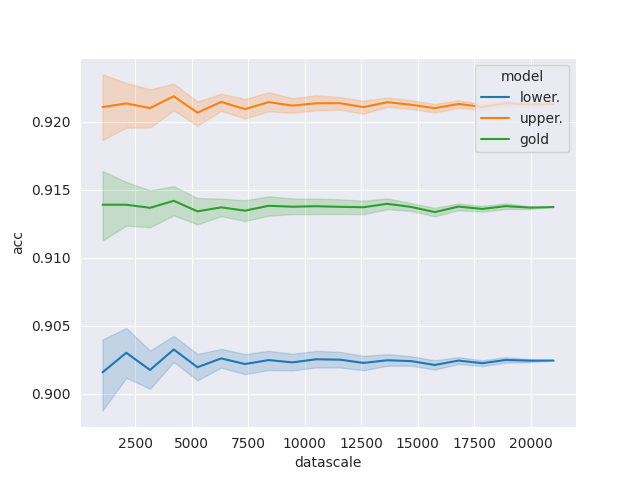}
        \caption{COGS}
        \label{fig:subfig_c}
    \end{subfigure}
    \vspace{0.01cm}

    \begin{subfigure}{.3\textwidth}
        \centering
        \includegraphics[width=\linewidth]{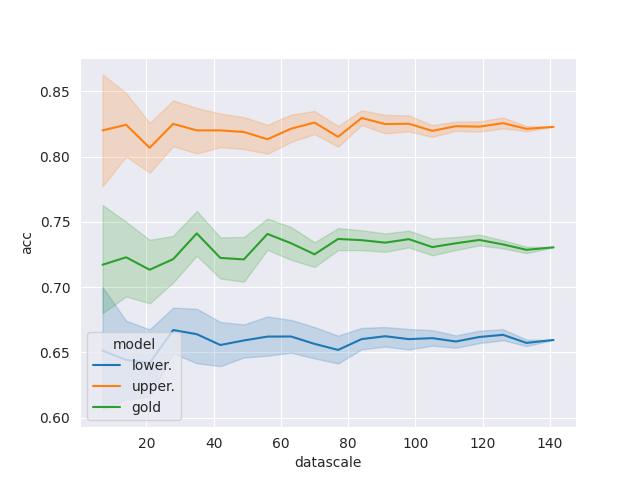}
        \caption{TOP}
        \label{fig:subfig_d}
    \end{subfigure}%
    \begin{subfigure}{.3\textwidth}
        \centering
        \includegraphics[width=\linewidth]{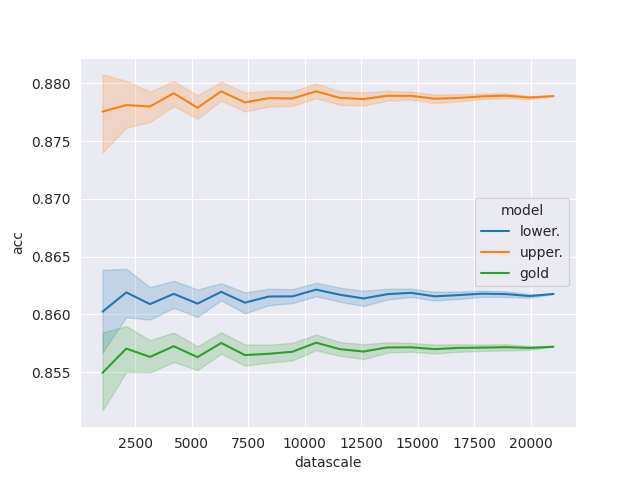}
        \caption{POS-COGS}
        \label{fig:subfig_e}
    \end{subfigure}%
    \begin{subfigure}{.3\textwidth}
        \centering
        \includegraphics[width=\linewidth]{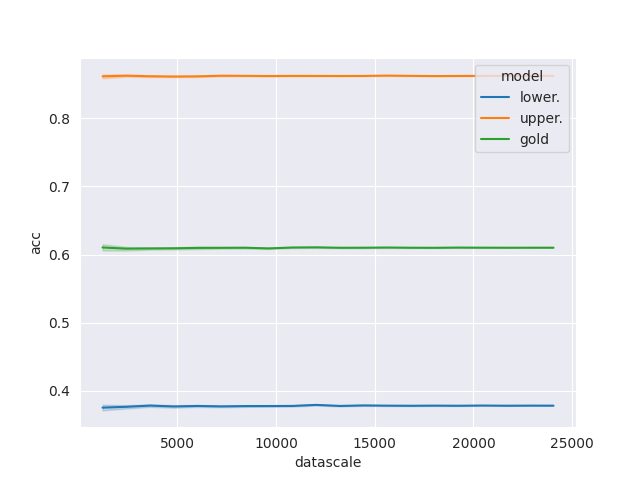}
        \caption{POS-Brown}
        \label{fig:subfig_f}
    \end{subfigure}

    \caption{Predicted accuracy w.r.t.\ the number of test-set instances. For each subset we randomly sample 50 times and show its confidence interval with 95\% confidence.}
    \label{fig:acc_vs_scale}
\end{figure*}

\section{Discussion} \label{sec:discussion}

\paragraph{Low performance on constituency parsing.} 
Our method predicts loose bounds on PTB parsing tasks and sometimes yields high AE scores. 
We conjecture that this is because the PTB training set contains many long output sequences (e.g.\ linearized parse trees), whose lengths are much larger than the maximal encoding length (e.g.\ 512) of our language model discriminators. 
Encoding sequences longer than observed during pretraining has been shown challenging for transformer-based language models \cite{dai-etal-2019-transformer}, which leads to an additional challenge for our discriminators. 
Nonetheless, the gold accuracy is still robustly between the predicted bounds.


\paragraph{The robustness of discriminators.} 
We have seen that our predicted upper and lower bounds can capture the gold accuracy. 
However, this may not be enough to show the robustness of our method, since we only evaluated it on one overall test set for each parser, while previous works \cite{garg2022leveraging} collect multiple test sets for each classifier.  
To investigate the robustness of our method, we create multiple test sets by randomly sampling subsets from the original test set and plot the accuracy of T5 discriminators on OOD test sets in Figure \ref{fig:acc_vs_scale}. See Appendix \ref{appendix:subsets} for full results of both T5 and {RoBERTa} discriminators.

According to the results, we can observe that our predicted bounds robustly capture the gold accuracy with regard to different sizes of randomly sampled test sets.   
On COGS, POS-COGS and TOP, a small test set gives a large confidence interval. 
We consider this is because their test sets contain some extremely difficult examples for the parser, which could result in a challenging subset and yield a low accuracy. 
Despite this, our discriminators capture the difficulty of such challenging subsets and shares similar confidence intervals as the gold accuracy. 

\paragraph{Training models to predict another model's accuracy.}
The main idea of our method is to train one model (i.e.\ discriminator) to predict a base model's accuracy (i.e.\ parser).
Previous work estimates model accuracy using the parser's confidence scores, which can be misleading in OOD tasks due to overconfidence, as shown in Section \ref{sec:results}.
On the other hand, \citet{kim-etal-2021-improving} find that despite the parser's poor performance on OOD data, categorizing the parsing task into a classification task enables their classifier to generalize to OOD data to some extents.  
Hence, we directly train a classifier (i.e.\ discriminator) to assess correctness of the prediction.
Training discriminators to evaluate another model's prediction has been studied in various fields, including adversarial learning \cite{goodfellow2014generative}, error detection \cite{chen-etal-2023-error} and reranking \cite{yin-neubig-2019-reranking}, where discriminators are used to improve the base model's accuracy.
Our work differs in using discriminators to predict bounds of the base model's accuracy. 


Despite the impressive performance of our discriminators, 
training them relies on incorrect parser outputs (i.e.\ errors) on the parser's training set.
This raises concerns about the generalization of discriminators to errors made in OOD data by the parser.
Our results empirically show that discriminators can generalize to such errors, 
but further investigation is needed to understand the specific scenarios where this method fails and the underlying reasons.
Since the discriminators' performance still lags behind on particular tasks, it is worth exploring discriminators' confidence scores to improve the estimated accuracy. We leave this for future study.


\section{Conclusion}

We propose a method to predict \textit{upper and lower bounds} for the accuracy of a model on unlabeled and possibly OOD data. 
To do this, we first train multiple correctness discriminators by finetuning pretrained language models, and then ensemble discriminator predictions through a special voting mechanism. 
Our experiments show that our predicted bounds reliably capture gold accuracy across a variety of in-distribution and out-of-distribution tasks including semantic parsing, tagging and constituency parsing tasks, and the upper and lower bounds are usually tight.
Although our method is not specifically designed for point estimation, simple heuristics (e.g.\ using the mean of bounds as estimated accuracy) based on our method can substantially outperform previous methods, which indicates the effectiveness of our method.

For the future, we will explore the use of our discriminators to improve model performance on tasks evaluated in this paper. 
For example, given unlabeled OOD sentences and a parser, our lower bound can be used to detect instances with a high \textit{Correct-Precision}, as training data to improve the parser. 
It would be interesting to expand our prediction of hard upper and lower bounds to
a Bayesian model that predicts a probability distribution over accuracies.
Finally, it would be useful to extend our method to predicting accuracy in terms of other metrics
 (e.g.\ parsing f-score or SMATCH).


 \section*{Limitations}

 In order to train the discriminators, we extract negative training instances from the beam of a partially finetuned T5 model. Our method is therefore not applicable in situations where we cannot easily access the beam, e.g.\ when trying to estimate the accuracy of a closed language model.

 Training the discriminators incurs a computational overhead, compared to training only the parser. In the experiments reported above, with five discriminators per ensemble, the training time is increased roughly by a factor of ten. However, once an ensemble has been trained, it can be applied across many unlabeled test sets for the same task.

 Finally, the discriminator ensembles in our approach currently vote only at the level of entire test instances, which means that we can only use instance-level evaluation measures such as exact match. As we discussed in the conclusion, extending our approach to other evaluation measures seems like a very useful topic for future research.

\section*{Acknowledgements}
We appreciate Mareike Hartmann, Megan Dare, Blerta Veseli and Xudong Hong for their insightful feedback to this paper.
This work was supported by the Deutsche Forschungsgemeinschaft (DFG) through the project KO 2916/2-2. 






\bibliography{anthology,custom}
\appendix

\section{Dataset details} \label{appendix:dataset}
We introduce details of our used datasets here. Statistics of datasets are reported in Table \ref{tab:dataset_statistics}.
The license of datasets are reported in Table \ref{tab:dataset_license}.
\begin{table*}
    \centering
    \footnotesize
    \begin{tabular}{lccccccccc}
    \toprule
    Dataset & Split & \# train & \# dev. & \# test & \# gen & Vocab. size & Train len. & Test len. & Gen len. \\
    \midrule
    COGS & - & 24155 & 3000 & 3000 & 21000 & 809 & $22 / 48$ &$19/40 $ & $61 / 144$ \\
    \multirow{2}{*}{CFQ} & MCD1 & 95743 & 11968 & 11968 & - & 171 & $29/ 133$ & $30 / 103$ & - \\
    & MCD2 & 95743 & 11968 & 11968 & - & 171 & $29/ 133$ & $30 / 103$ & - \\
    TOP & weather & 84372 & - & 484 & - & 29462 & $51/ 82$ & $21 / 60$ & - \\
    AMR 2 & - & 36251 & 1368 & 1371 & - & 79651 & $216 / 615$ & $159/583$ & - \\
    POS-WSJ & - & 39832 & 1700 & 2416 & - & 46378 & $141/141$ & $67/67$ & - \\
    POS-Brown & - & - & - & 24243 & - & 29564 & - & $172/172$ & -\\ 
    POS-COGS & - & 24155 & 3000 & 3000 & 21000 & 753 & $22/21$ & $22/21$ & $61/60$ \\
    Syn-WSJ & - & 39832 & 1700 & 2416 & - & 46404 & $141/903$ & $67/429$ & - \\
    Syn-Brown & - & - & - & 24243 & - & 29596 & - & $172/1135$ & - \\
    \bottomrule
    \end{tabular}
    \caption{Statistics for all our datasets. \# denotes the number of instances in the dataset. Vocab.size denotes the size of vocabulary for the dataset, which consists of input tokens and output tokens. Train.len denotes the maximum length of the input tokens and output tokens in the train set. Test.len and Gen.len denote the maximum length in the test and generalization set.}
    \label{tab:dataset_statistics}
\end{table*}

\paragraph{COGS} \cite{kim-linzen-2020-cogs} is a synthetic English semantic parsing task. The task input is a sentence and the output is a logical form (e.g.\ \textit{The baby on a tray in the house screamed.} $\to$ \lform{ scream(agent=*baby(nmod.on= tray(nmod.in=*house)))}). It provides a training set generated by a probabilistic context-free grammar (PCFG) and a OOD test set with 21-typed data, which are generated by different PCFGs to test the different generalization abilities of the parser.

\paragraph{POS-COGS} \cite{yao-koller-2022-structural} is a synthetic English part-of-speech tagging task generated based on COGS. The task input is a sentence and the output is the POS tag sequence (e.g.\ \textit{The baby on a tray in the house screamed.} $\to$ \lform{Det N P Det N P Det N V}). POS-COGS shares the same split of train and test sets as COGS. 

\paragraph{CFQ} \cite{keysers-etal-2020-measuring} is a synthetic English semantic parsing task. The task input is a sentence and the output is a SPARQL query (e.g.\ \textit{Did M0 ' s writer write M1 and M2} $\to$ \lform{SELECT count(*) WHERE \{?x0 film.writer.film M0...\}}). We use the MCD1 and MCD2 splits of CFQ, where the test set is designed to compositionally diverge from the training set but share similar atom distributions.

\paragraph{TOPv2} \cite{chen-etal-2020-low} is a natural English semantic parsing task. The task input is a sentence and the output is a hierarchical semantic representation \cite{gupta-etal-2018-semantic-parsing} (e.g.\ \textit{Will there be snowfall this week?} $\to$ \lform{[in:get\_weather will there be [sl:weather\_attribute snowfall] [sl:date\_time this week] ?]}). The TOPv2 training set consists of data from multiple domains including two low-resource domains (e.g.\ \textit{reminder} and \textit{weather}), and the test set consists of data from the two domains to test low-resource domain adaptation ability of the parser. We focused on the\textit{weather} domain in our experiments.

\paragraph{AMR 2.0} \cite{banarescu-etal-2013-abstract} is an English semantic parsing task. 
The input is a sentence and output is an abstract meaning representation (e.g.\ \textit{I will stick around until the end}$\to$\lform{(stick-around-03 :ARG0(i) :time(until :op1(end-01)))}). 

\paragraph{Penn Treebank 3 (PTB)} \cite{marcus-etal-1993-building} is an English constituency parsing task. The input is a sentence and the output is the constituency parse tree (e.g.\ \textit{Vice President} $\to$ \lform{(TOP(NP(NNP Vice)(NNP President)))}).

\begin{table}
    \centering
    \footnotesize
    \begin{tabular}{lc}
    \toprule
    Dataset & License \\
    \midrule
    COGS & MIT \\
    CFQ & CC-BY \\
    TOP & CC-BY-SA \\
    AMR 2 & LDC \\
    PTB & LDC \\
    POS-COGS & MIT \\
    \bottomrule
    \end{tabular}
    \caption{Licenses for used datasets.}
    \label{tab:dataset_license}
\end{table}

\section{Training details} \label{appendix:training}
\subsection{Hyperparameters}
\paragraph{Parser.} We finetune \textit{t5-base}\footnote{\url{https://huggingface.co/t5-base}} (220M parameters) as our parser for all tasks. 
We use Adam \cite{DBLP:journals/corr/KingmaB14} as the optimizer.
For most tasks, the learning rate is set to 1e-5. For CFQ, AMR, PennTreebank tasks, the learning rate is set to 1e-4 to make the training faster.
For tasks that provide a development set, early stopping is used and the best checkpoint is selected based on the evaluation metrics on the development set. Otherwise, the checkpoint at the end of training is used to report results.
For AMR, the evaluation metric is Smatch F1 score. 
For syntactic parsing, the evaluation metric is EVALB F1 score\footnote{\url{https://nlp.cs.nyu.edu/evalb/}}.
For other tasks, exact match accuracy is used as the evaluation metric. 
We use weight decay 1e-3 for all datasets.
No learning rate scheduler is used for all experiments.
During evaluation, we use beam search with beam size 4.

\paragraph{Discriminator.} 
We finetune \textit{t5-base} (220M parameters), \textit{roberta-base} (125M parameters) and \textit{Vicuna-7b-v1.5}\footnote{\url{https://huggingface.co/lmsys/vicuna-7b-v1.5}} (7B parameters) as our discriminators.
To collect training data, we use the first 5 checkpoints of the parser and validate them on the parser's training set. 
We select negative examples from beam predictions of these checkpoints as the training data of the discriminator. 
If a task provides an in-distribution development set for the parser, we use the same method to create the development set for the discriminator. 

For T5, we follow the same hyperparameter settings described for T5 parser. 
For RoBERTa, we adopt learning rate 1e-5 for all tasks except PTB. On PTB tasks, the learning rate is set to 5e-5.  
For both T5 and RoBERTa, we validate the AUC score on the development set to select the best checkpoint of the discriminator when a development set is available. 
Otherwise, we train the discriminator until its training loss converges with fixed steps.
Note that although CFQ provides an out-of-distribution development set, we did not use it since we assume we do not have the access to the OOD data.

We use QLoRA \cite{dettmers2023qlora} to finetune Vicuna discriminators. 
For datasets except POS-COGS, the learning rate is set to 3e-4. For POS-COGS dataset, the learning rate is set to 2e-5. 
We use linear scheduler with warmup as our learning rate scheduler. 
We use weight decay 1e-3 for all datasets.
For LoRA, we set the rank value to 8, the alpha value to 32 and the dropout value to 0.1.

\begin{table*}[htb!]\centering
\footnotesize
\begin{tabularx}{\linewidth}{XXXXXXXXXXX}
\toprule
& \multicolumn{8}{c}{OOD} & \multicolumn{2}{c}{ID} \\
\cmidrule(lr){2-9}
\cmidrule(lr){10-11}
&\multicolumn{2}{c}{MCD1} &\multicolumn{2}{c}{MCD2} &\multicolumn{2}{c}{COGS} &\multicolumn{2}{c}{TOP} &\multicolumn{2}{c}{AMR 2.0} \\\cmidrule{2-11}
&Acc &AE $\downarrow$ &Acc &AE $\downarrow$&Acc &AE $\downarrow$&Acc &AE $\downarrow$ &Acc &AE $\downarrow$\\\midrule
\multicolumn{11}{l}{\textit{Ours (Vicuna)}} \\
\midrule
{Mean}\textsubscript{discrim} & 56.4 &1.4 &23.6 &0.7 &93.0 &1.6 &76.2 &3.2 & 18.3& 4.0 \\
Upper. &\textcolor{mdgreen}{60.0} &- &\textcolor{mdgreen}{28.1} &- &\textcolor{mdgreen}{94.7} &- &\textcolor{mdgreen}{87.9} &- &\textcolor{mdgreen}{28.0} &- \\
Lower. &\textcolor{mdgreen}{51.5} &- &\textcolor{mdgreen}{18.4} &- &\textcolor{mdgreen}{87.8} &- &\textcolor{mdgreen}{57.4 }&- &\textcolor{mdgreen}{4.2} &- \\
{Mean}\textsubscript{bounds} &55.8 &{2.0} &23.3 &{0.4} &91.3 &{0.2} &72.7 &{0.3} &16.1 &1.8 \\\midrule
Gold &57.8 &0.0 &22.9 &0.0 &91.4 &0.0 &73.0 &0.0 &14.3 &0.0 \\
\bottomrule
\end{tabularx}
\caption{Predicted test-set accuracy with Vicuna-based discriminators on semantic parsing tasks. }
\label{tab:app:sem_acc}
\end{table*}

\begin{table}[!htbp]
    \centering
    \footnotesize
    \begin{tabular}{lccc}
    \toprule
          & \multicolumn{3}{c}{Time (hours)}  \\
         Dataset & T5 & Roberta & Vicuna\\
         \midrule
         MCD1 &  10 & 8 & 12 \\
         MCD2 &  10 & 8 & 12 \\
         COGS & 3 & 3 & 5 \\
         Top &2 & 5 & 2 \\
         AMR & 16 & 18 & 50\\
         POS-WSJ & 16 & 16 &100\\
         POS-COGS & 2 & 3 & 5\\
         Syn-WSj & 15 & 10 & 100\\
         
         \bottomrule
    \end{tabular}
    \caption{Training time for our model on each dataset (1 run) in our experiments.}
    \label{tab:traintime}
\end{table}

\subsection{Other details}
We use AllenNLP \cite{gardner-etal-2018-allennlp} to implement T5 and RoBERTa finetuning and Huggingface \cite{wolf-etal-2020-transformers} to implement Vicuna finetuning. Experiments are run on Tesla
A100 GPU cards (80GB). Table \ref{tab:traintime} shows the training time cost to train a single discriminator on one GPU. 

\begin{table}[!htb]\centering
\scriptsize
\begin{tabularx}{\linewidth}{p{0.45cm}p{1.3cm}XXXXXX}\toprule
& &\multicolumn{2}{c}{Single} &\multicolumn{2}{c}{Upperbound} &\multicolumn{2}{c}{Lowerbound} \\
\cmidrule(lr){3-4}
\cmidrule(lr){5-6}
\cmidrule(lr){7-8}
& &CR &IR &CR &IR &CR &IR \\
\midrule
\multirow{7}{*}{OOD} &MCD1 &94.3 &96.7 &98.7 &93.0 &88.4 &98.9 \\
&MCD2 &81.3 &94.7 &95.9 &92.0 &68.1 &96.4 \\
&COGS &98.8 &78.0 &99.9 &60.2 &95.6 &95.5 \\
&TOP &82.5 &71.1 &97.1 &36.8 &72.8 &84.2 \\
&POS-Brown &68.6& 60.5& 92.1& 26.5& 24.7&94.6 \\
&POS-COGS &97.9 &86.4 &98.9 &72.6 &96.3 &91.5 \\
&Syn-Brown &56.9& 59.3& 88.0& 27.5& 25.3& 75.0\\
\midrule
\multirow{3}{*}{ID} &AMR 2.0 &51.3 & 89.9 &57.5 &76.9 & 26.4 & 99.5 \\
&POS-WSJ &70.4 &58.2 &91.2 &26.6 &33.6 &93.0 \\
&Syn-WSJ &63.9& 52.4& 91.7& 23.8& 32.2&68.1 \\
\bottomrule
\end{tabularx}
\caption{Correct recall and incorrect recall scores of Vicuna-based discriminators.}
\label{tab: app:recall}
\end{table}

\begin{table}[htb!]\centering
\footnotesize
\begin{tabularx}{\linewidth}{p{1.3cm}XXXXXX}\toprule
& \multicolumn{4}{c}{OOD} & \multicolumn{2}{c}{ID} \\
\cmidrule(lr){2-5}
\cmidrule(lr){6-7}
&\multicolumn{2}{c}{POS-Brown} &\multicolumn{2}{c}{POS-COGS} &\multicolumn{2}{c}{POS-WSJ} \\\cmidrule{2-7}
&Acc &AE &Acc &AE &Acc &AE \\
\midrule
\multicolumn{7}{l}{\textit{Ours (Vicuna)}} \\
\midrule
{Mean}\textsubscript{discrim} & 51.5 &9.5 &86.8 &1.1 &55.2 &10.1 \\
Single & 57.3& 3.7&85.3 &{0.4} &60.5 &4.8 \\
Upper. & \textcolor{mdgreen}{84.8}& -&\textcolor{mdgreen}{86.6} &- & \textcolor{mdgreen}{85.0}& -\\
Lower. & \textcolor{mdgreen}{17.2}& -&\textcolor{mdgreen}{83.8} &- & \textcolor{mdgreen}{24.4}& -\\
{Mean}\textsubscript{bounds} & 51.0& 10.0&85.2 &0.5 & 54.7& 10.6\\
\midrule
Gold &61.0 &0.0 &85.7 &0.0 &65.3 &0.0 \\
\bottomrule
\end{tabularx}
\caption{Predicted accuracy on POS tagging tasks.}
\label{tab:app:pos_acc}
\end{table}

\begin{table}[htb!]\centering
\footnotesize
\begin{tabularx}{\linewidth}{XXXXX}\toprule
&\multicolumn{2}{c}{OOD} &\multicolumn{2}{c}{ID} \\
\cmidrule(lr){2-3}
\cmidrule(lr){4-5}
&\multicolumn{2}{c}{Syn-Brown} &\multicolumn{2}{c}{Syn-WSJ} \\\cmidrule{2-5}
&Acc &AE &Acc &AE \\\midrule
\multicolumn{5}{l}{\textit{Ours (Vicuna)}}\\\midrule
{Mean}\textsubscript{discrim} &51.5 &17.7 & 56.7& 19.1\\
Upper.&\textcolor{mdgreen}{77.8} &- & \textcolor{mdgreen}{82.1}& -\\
Lower. &\textcolor{mdgreen}{25.1} &- & \textcolor{mdgreen}{31.2}& -\\
{Mean}\textsubscript{bounds} &51.5 &17.7 & 56.7& 19.1\\
\midrule
Gold &33.8 &0.0 &37.6 &0.0 \\
\bottomrule
\end{tabularx}
\caption{Predicted accuracy on parsing tasks.}
\label{tab:app:syn_acc}
\end{table}
\section{Results of Vicuna-based discriminators}\label{appendix:architecture}


Here we report the results of Vicuna discriminators in Table \ref{tab: app:recall}.
In this setting, we finetune \textit{Vicuna-7B} on the same datasets we used in Section \ref{sec:datasets}. 


Similar to the observation in Section \ref{sec:experiments}, Vicuna-based discriminators achieve high recall scores on most datasets, indicating that they can still make correct judgements for most instances. 
We also report estimation errors of this discriminator in Table \ref{tab:app:sem_acc}, \ref{tab:app:pos_acc}, \ref{tab:app:syn_acc}.
According to the results, we can observe that Vicuna-based discriminators still achieve very low estimation errors across different datasets. 
The predicted upper and lower bounds captures the gold accuracy on all datasets. 
These results are consistent with our observations when using T5 and RoBERTa discriminators, which suggests that our method is robust with regard to different discriminator architectures.





\section{Results of discriminators on subsets of test sets} \label{appendix:subsets}

\begin{figure*}[tb!]
    \centering
    \begin{subfigure}{.24\textwidth}
        \centering
        \includegraphics[width=\linewidth]{figures/acc_vs_scale/mcd1_acc_wrt_datascale.png}
        \caption{MCD1}
    \end{subfigure}%
    \begin{subfigure}{.24\textwidth}
        \centering
        \includegraphics[width=\linewidth]{figures/acc_vs_scale/mcd2_acc_wrt_datascale.png}
        \caption{MCD2}
    \end{subfigure}%
    \begin{subfigure}{.24\textwidth}
        \centering
        \includegraphics[width=\linewidth]{figures/acc_vs_scale/cogs_acc_wrt_datascale.png}
        \caption{COGS}
    \end{subfigure}
    \begin{subfigure}{.24\textwidth}
        \centering
        \includegraphics[width=\linewidth]{figures/acc_vs_scale/top_acc_wrt_datascale.png}
        \caption{TOP}
    \end{subfigure}%
    
    \begin{subfigure}{.24\textwidth}
        \centering
        \includegraphics[width=\linewidth]{figures/acc_vs_scale/pos_cogs_acc_wrt_datascale.png}
        \caption{POS-COGS}
    \end{subfigure}%
    \begin{subfigure}{.24\textwidth}
        \centering
        \includegraphics[width=\linewidth]{figures/acc_vs_scale/pos_brown_acc_wrt_datascale.png}
        \caption{POS-Brown}
    \end{subfigure}
    \vspace{0.01cm}
    \begin{subfigure}{.24\textwidth}
        \centering
        \includegraphics[width=\linewidth]{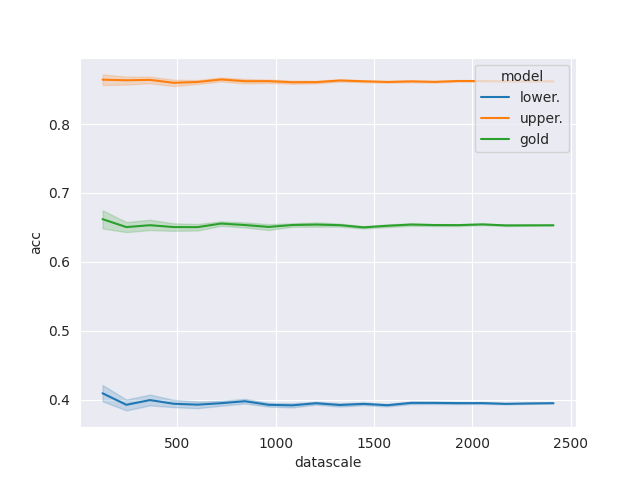}
        \caption{POS-WSJ}
    \end{subfigure}%
    \begin{subfigure}{.24\textwidth}
        \centering
        \includegraphics[width=\linewidth]{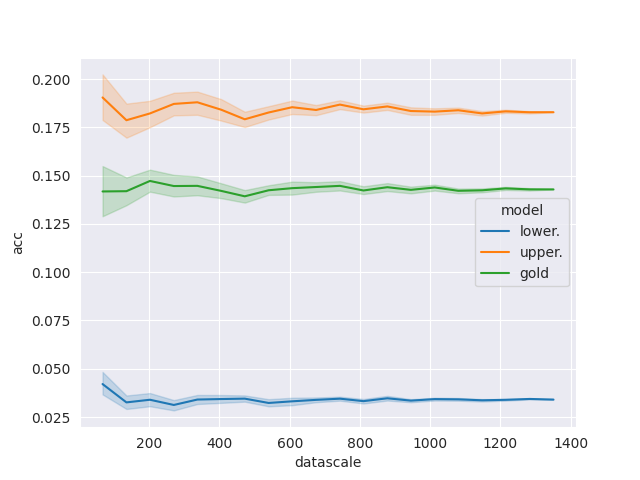}
        \caption{POS-COGS}
    \end{subfigure}%
    
    \begin{subfigure}{.24\textwidth}
        \centering
        \includegraphics[width=\linewidth]{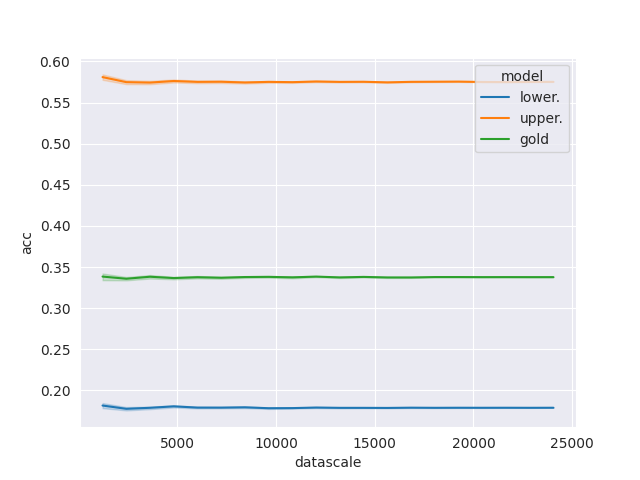}
        \caption{Syn-Brown}
    \end{subfigure}
    \begin{subfigure}{.24\textwidth}
        \centering
        \includegraphics[width=\linewidth]{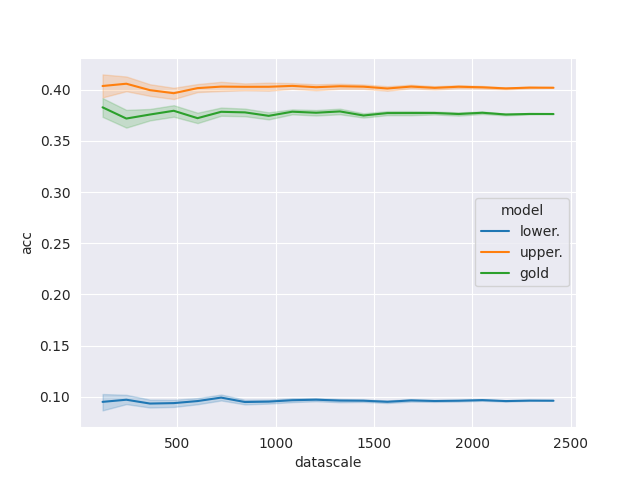}
        \caption{Syn-WSJ}
    \end{subfigure}%

    \caption{Predicted accuracy by T5 discriminators w.r.t.\ the number of test-set instances. For each subset we randomly sample 50 times and show its confidence interval with 95\% confidence.}
    \label{fig:t5_acc_vs_scale}
\end{figure*}

\begin{figure*}[tb!]
    \centering
    \begin{subfigure}{.24\textwidth}
        \centering
        \includegraphics[width=\linewidth]{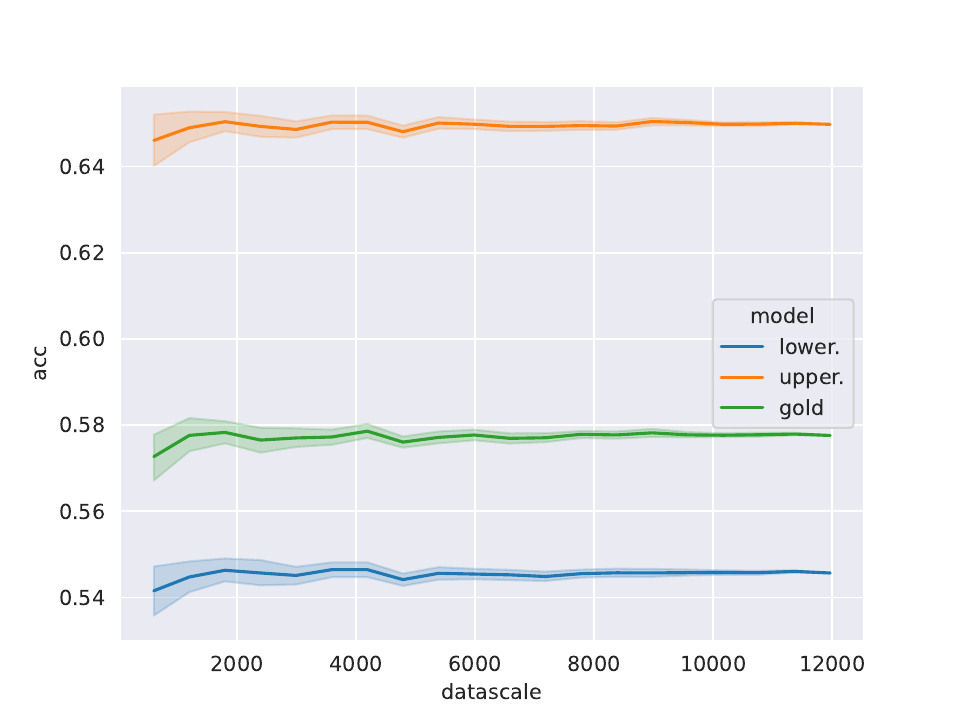}
        \caption{MCD1}
    \end{subfigure}%
    \begin{subfigure}{.24\textwidth}
        \centering
        \includegraphics[width=\linewidth]{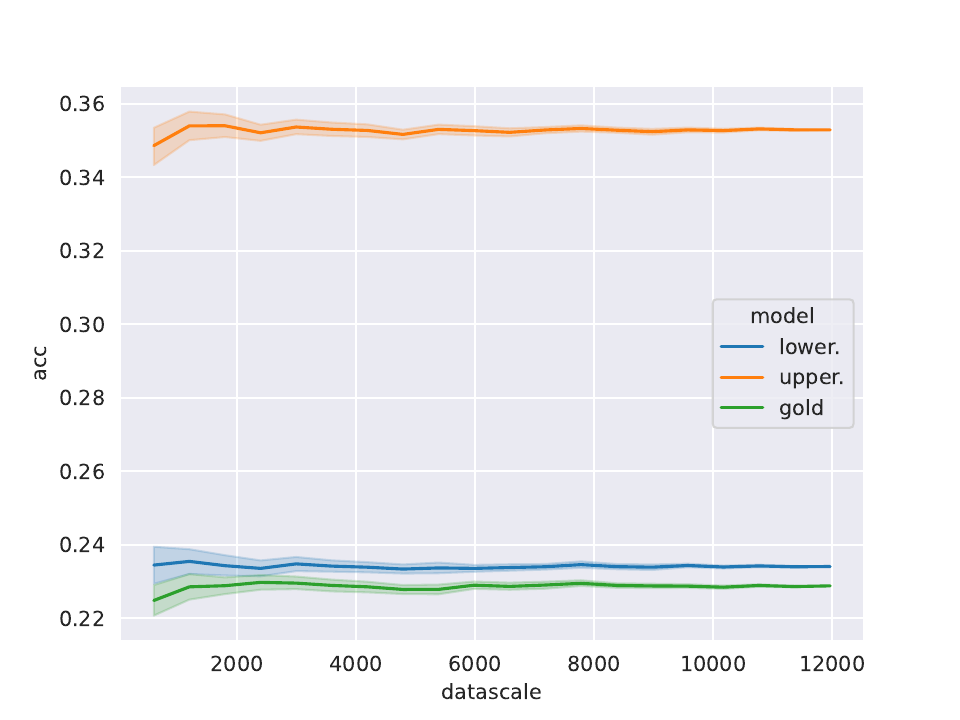}
        \caption{MCD2}
    \end{subfigure}%
    \begin{subfigure}{.24\textwidth}
        \centering
        \includegraphics[width=\linewidth]{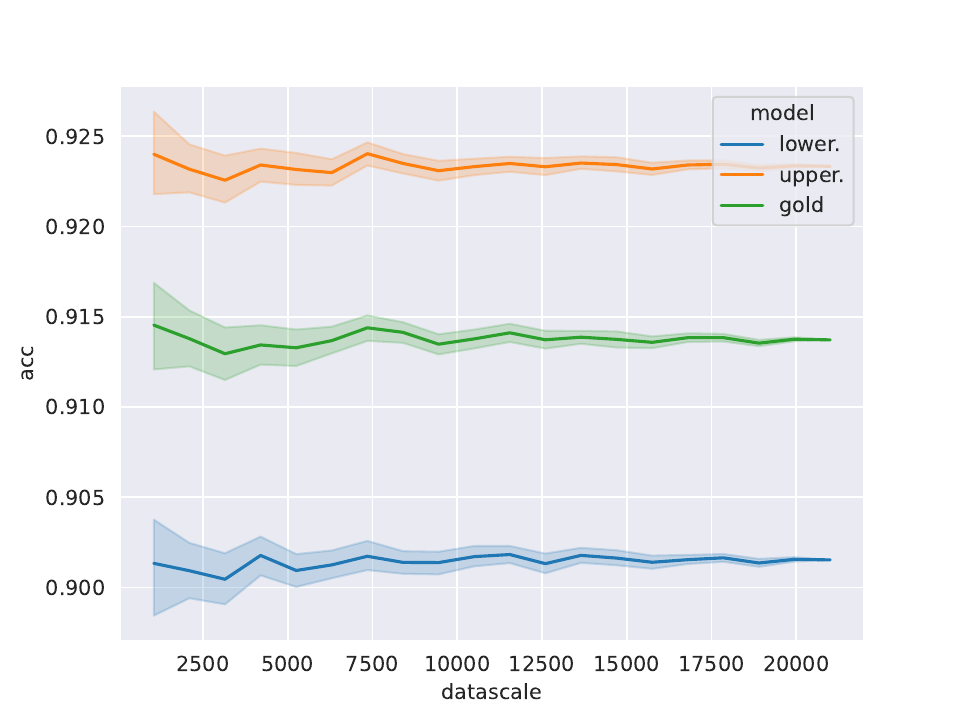}
        \caption{COGS}
    \end{subfigure}
    \begin{subfigure}{.24\textwidth}
        \centering
        \includegraphics[width=\linewidth]{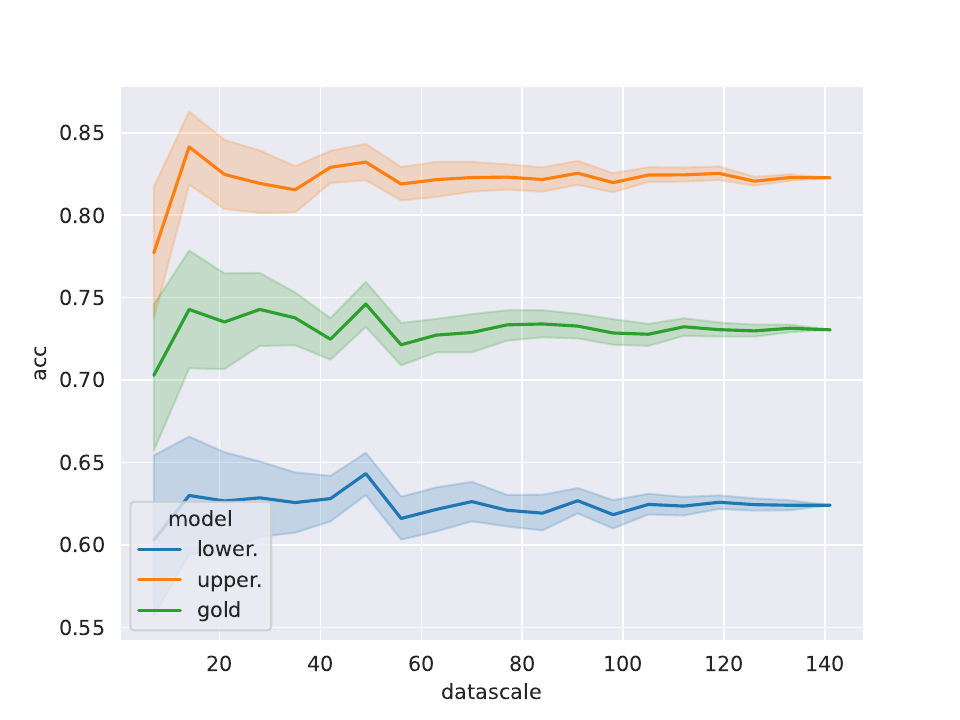}
        \caption{TOP}
    \end{subfigure}%
    
    \begin{subfigure}{.24\textwidth}
        \centering
        \includegraphics[width=\linewidth]{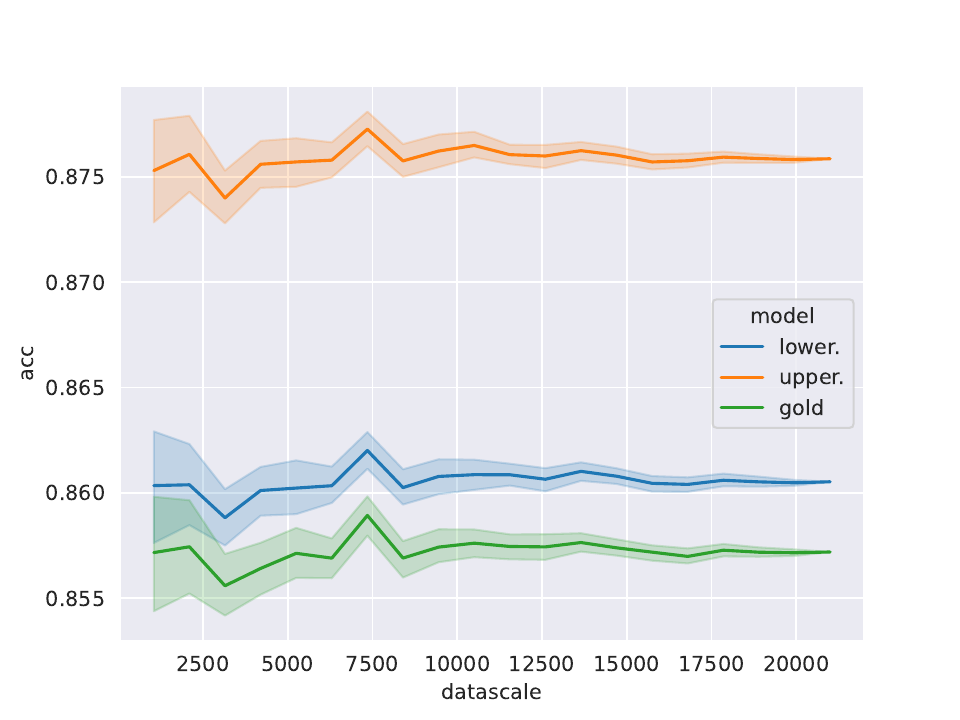}
        \caption{POS-COGS}
    \end{subfigure}%
    \begin{subfigure}{.24\textwidth}
        \centering
        \includegraphics[width=\linewidth]{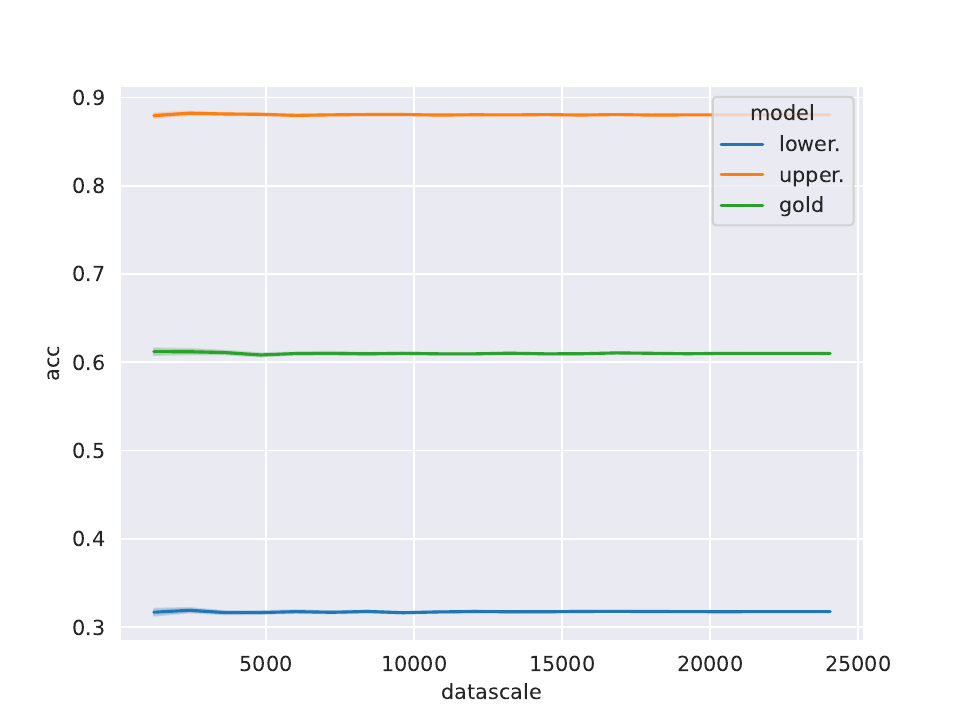}
        \caption{POS-Brown}
    \end{subfigure}
    \vspace{0.01cm}
    \begin{subfigure}{.24\textwidth}
        \centering
        \includegraphics[width=\linewidth]{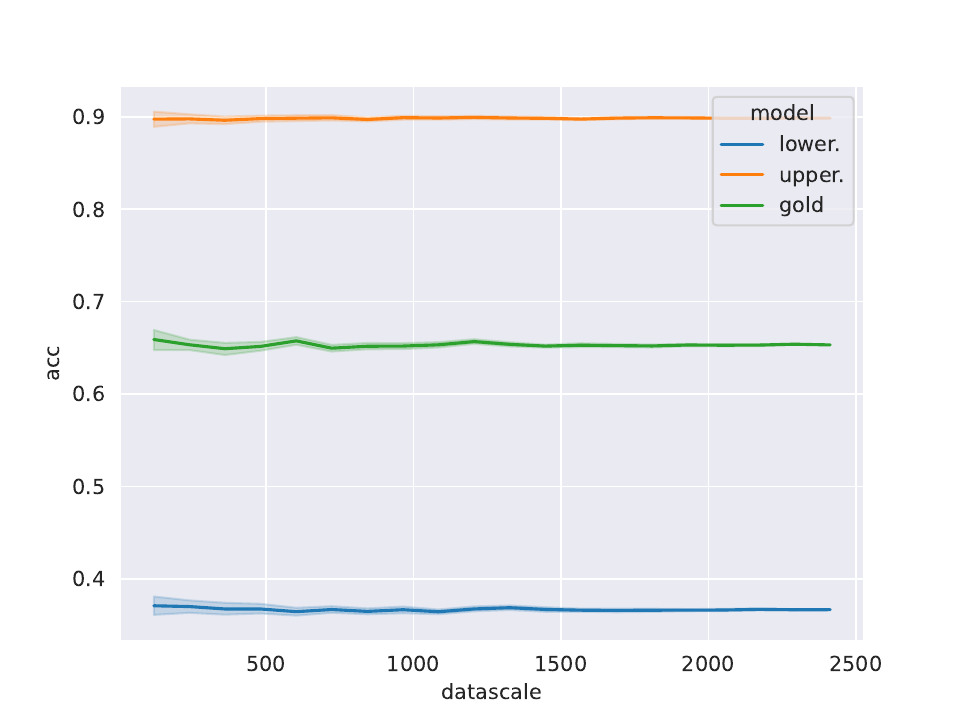}
        \caption{POS-WSJ}
    \end{subfigure}%
    \begin{subfigure}{.24\textwidth}
        \centering
        \includegraphics[width=\linewidth]{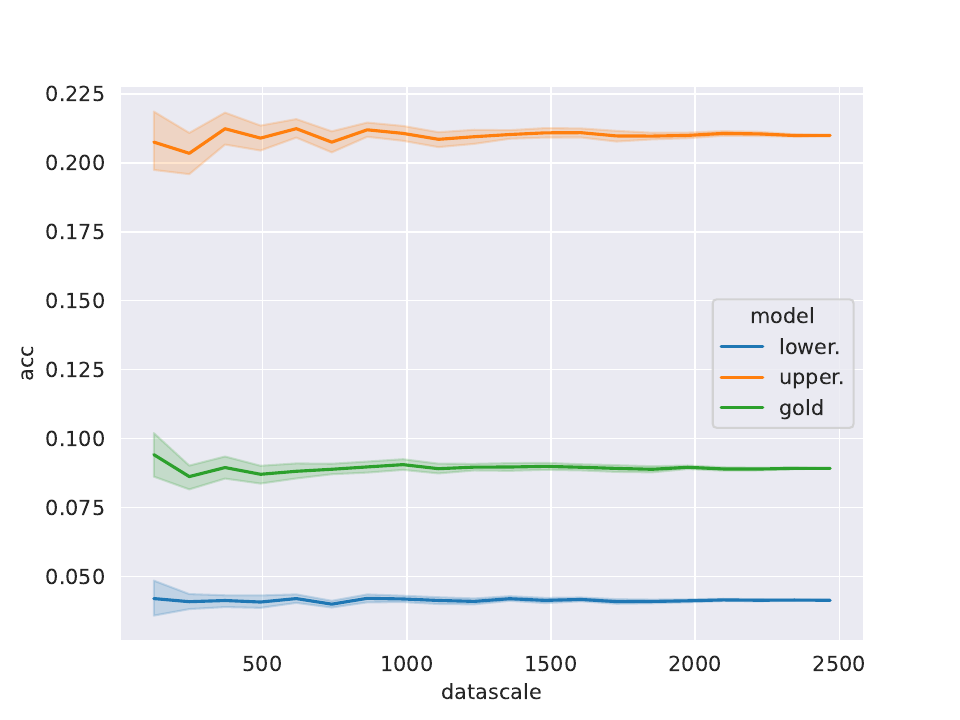}
        \caption{POS-COGS}
    \end{subfigure}%
    
    \begin{subfigure}{.24\textwidth}
        \centering
        \includegraphics[width=\linewidth]{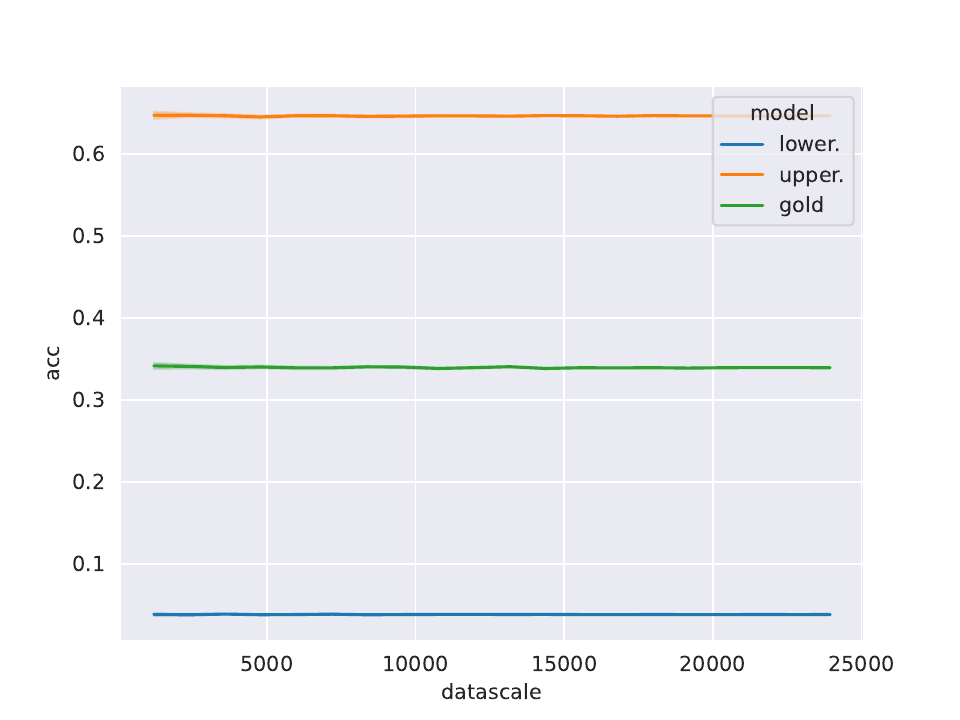}
        \caption{Syn-Brown}
        \label{fig:subfig_i}
    \end{subfigure}
    \begin{subfigure}{.24\textwidth}
        \centering
        \includegraphics[width=\linewidth]{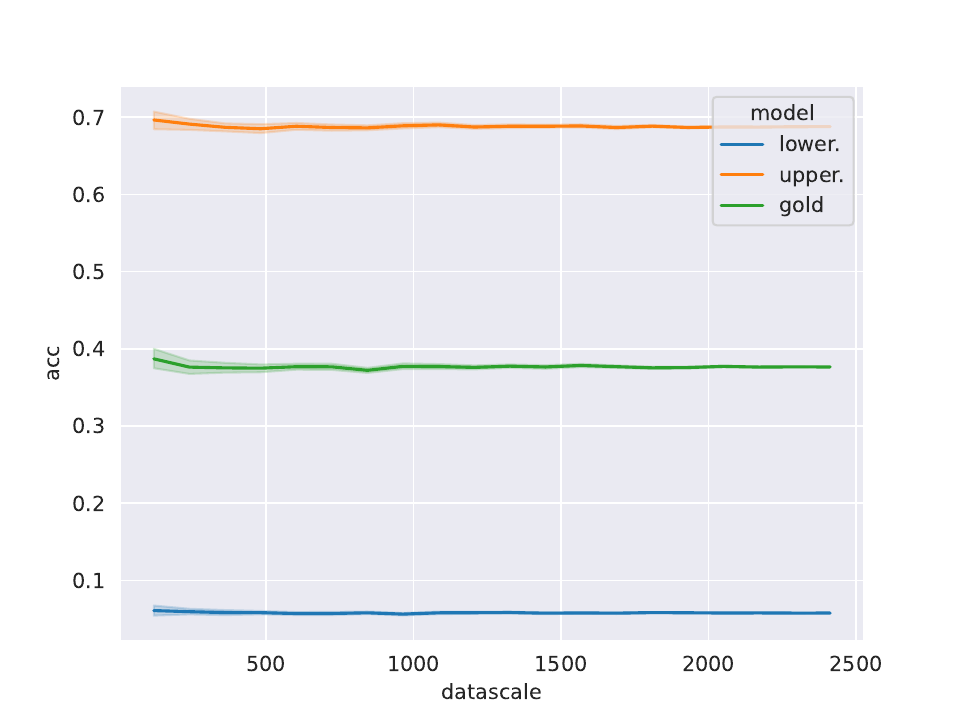}
        \caption{Syn-WSJ}
        \label{fig:subfig_g}
    \end{subfigure}%

    \caption{Predicted accuracy by RoBERTa discriminators w.r.t.\ the number of test-set instances. For each subset we randomly sample 50 times and show its confidence interval with 95\% confidence.}
    \label{fig:roberta_acc_vs_scale}
\end{figure*}

We report results of T5 and RoBERTa discriminators on different subsets of the test set in Figure \ref{fig:t5_acc_vs_scale}, \ref{fig:roberta_acc_vs_scale}.
Both discriminators predict lower and upper bounds that capture the gold accuracy robustly, which is consistent with our observation in Section \ref{sec:discussion}.

\end{document}